# Identity Documents Recognition and Detection using Semantic Segmentation with Convolutional Neural Network

Mykola Kozlenko[a], Volodymyr Sendetskyi[b], Oleksiy Simkiv[b], Nazar Savchenko[b], and Andy Bosyi[b]

[a] *Vasyl Stefanyk Precarpathian National University, 57 Shevchenko str., Ivano Frankivsk, 76018, Ukraine*
[b] *MindCraft AI LLC, 19 Lisna str., Lviv, 79010, Ukraine*

**Abstract**
Object recognition and detection are well-studied problems with a developed set of almost standard solutions. Identity documents recognition, classification, detection, and localization are the tasks required in a number of applications, particularly, in physical access control security systems at critical infrastructure premises. In this paper, we propose the new original architecture of a model based on an artificial convolutional neural network and semantic segmentation approach for the recognition and detection of identity documents in images. The challenge with the processing of such images is the limited computational performance and the limited amount of memory when such an application is running on industrial one-board microcomputer hardware. The aim of this research is to prove the feasibility of the proposed technique and to obtain quality metrics. The methodology of the research is to evaluate the deep learning detection model trained on the mobile identity document video dataset. The dataset contains five hundred video clips for fifty different identity document types. The numerical results from simulations are used to evaluate the quality metrics. We present the results as accuracy versus threshold of the intersection over union value. The paper reports an accuracy above 0.75 for the intersection over union (IoU) threshold value of 0.8. Besides, we assessed the size of the model and proved the feasibility of running the model on an industrial one-board microcomputer or smartphone hardware.

## 1. Introduction

Almost every organization today uses access control security systems. Usually, employees use special access cards. But there is a problem for guests or people who visit an object for the first time and do not have an access card. In this case, the identification of the person can be performed according to the data of any official identity document. Identification can be done by detecting a document in an image from a camera or scanner followed by extraction of text information.

Object recognition and detection are well-studied problems with a developed set of almost standard solutions. Identity documents recognition, classification, detection, and localization are very popular tasks in the computer vision area and are required in many security applications [1]. Nowadays there are some classical approaches to object detection: Viola-Jones object detection framework based on Haar features [2], scale-invariant feature transform [3], a histogram of oriented gradients [4], etc. Also, object detection algorithms are implemented in popular frameworks and libraries such as OpenCV and many others. There are many deep learning-based approaches as well [5]. In this paper, we propose a new neural network (NN) architecture and investigate the performance of the semantic segmentation-based approach for identity documents detection.

## 2. Related Work

In recent years, many successful approaches to object detection using deep learning were proposed. R-CNN solution was proposed first in [6]. Reference [7] presents the Fast R-CNN. The Faster R-CNN is reported in [8]. Also, the following are well-known and widely used approaches. Single Shot MultiBox Detector (SSD) [9] approach is based on a feed-forward convolutional network that produces a collection of bounding boxes and scores for the presence of object class instances. One of the most popular object detectors is the You Only Look Once (YOLO) detector [10]. YOLO sees the entire image during training and test time so it implicitly encodes contextual information about classes [10]. It outperforms all other detection methods, including R-CNN. There are also some other well-known methods: Single-Shot Refinement Neural Network for Object Detection (RefineDet) [11], Retina-Net [12], Deformable convolutional networks [13], and others. Reference [14] is devoted to identity document recognition in a video stream. The paper [15] studies the problem of image classification of identity documents composed of few textual information fields and complex backgrounds. The proposed approach simultaneously locates the document and recognizes the class. Paper [16] discusses the problem of simultaneous document type recognition and projective distortion parameter estimation for the images of identity documents. The problem of face detection on identity documents under unconstrained environments was sufficiently studied in [17]. In [18] it is proposed the original neural network architecture for the semantic image segmentation task contains layers calculating direct and transposed integral Fast Hough Transform operators.

## 3. Dataset

In this research, we use the Mobile Identity Document Video dataset (MIDV-500) [19]. It consists of 500 video clips for 50 different identity document types with ground truth. The dataset contains data on 17 types of ID cards, 14 types of passports, 13 types of driving licenses, and 6 other identity documents of various countries. Each captured frame had the same resolution of 1080 by 920 pixels. There are the following cases in the dataset: the document lies on the table with homogeneous background, the document lies on various keyboards, the document is held by a hand, the document is partially hidden, the background is stuffed with unrelated objects. Total counts of train and test samples are 10500 and 4500. Some instances of images are presented in Fig. 1. Fig. 1 also shows the detection results obtained using OpenCV (light green boundaries). There are several images in which this approach works well, such as the bottom-right picture in the figure. But for most of the images, we conclude that the conditions are very diverse. A simple image processing algorithm cannot cover all the variety of colors, lighting, shadows, blur, and other differences. We converted the data in our dataset into the following structure (refer to Fig. 2), where: the 'path' is the path to an image within the dataset, the 'x0', 'y0', 'x1', 'y1', 'x2', 'y2', 'x3', 'y3' are the ground truth coordinates of quadrilateral vertices of the document image, the 'part' is a number specifying a part of the dataset, the 'group' is the background used in the image.

The idea of the data import is simple: iterate over all the images, resize them, draw the ground truth in a blank image, and store them in the corresponding variables. Then, we can simply return a batch of a certain size.

## 4. Method and Model Design

The proposed architecture of the artificial convolutional neural network (CNN) is presented in Fig. 3. The idea behind this model is as follows: we downsample the input image to the size 8x8 while learning some features about most of the regions. Then, we pass those features to a few dense layers that make a decision on whether there is an identity document in the image and if yes, where it is located. Finally, we use that decision and features calculated in the downsampling part.

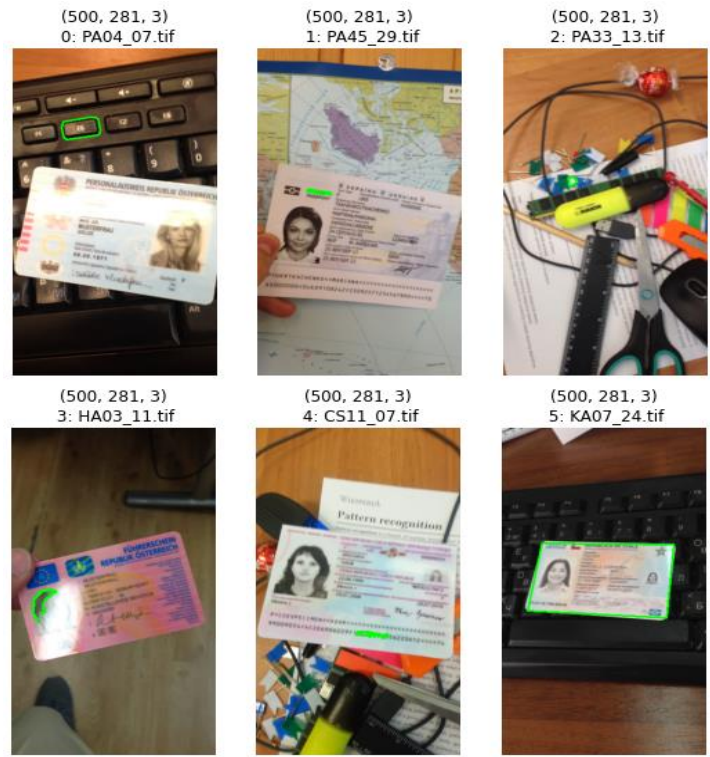

**Figure 1:** Examples of identity documents and backgrounds from the dataset

|  | path | x0 | y0 | x1 | y1 | x2 | y2 | x3 | y3 | part | group |
|---|---|---|---|---|---|---|---|---|---|---|---|
| 1 | 01_alb_id/images/CA/CA01_02.tif | 96 | 674 | 900 | 644 | 928 | 1141 | 121 | 1186 | 1 | CA |
| 2 | 01_alb_id/images/CA/CA01_14.tif | 186 | 772 | 865 | 753 | 887 | 1175 | 203 | 1208 | 1 | CA |
| 3 | 01_alb_id/images/CA/CA01_05.tif | 106 | 699 | 885 | 667 | 913 | 1155 | 130 | 1199 | 1 | CA |
| 4 | 01_alb_id/images/CA/CA01_27.tif | 80 | 657 | 904 | 623 | 933 | 1139 | 108 | 1186 | 1 | CA |
| 5 | 01_alb_id/images/CA/CA01_23.tif | 71 | 660 | 937 | 630 | 965 | 1166 | 100 | 1214 | 1 | CA |
| 6 | 01_alb_id/images/CA/CA01_03.tif | 95 | 674 | 898 | 646 | 925 | 1143 | 119 | 1189 | 1 | CA |
| 7 | 01_alb_id/images/CA/CA01_07.tif | 138 | 731 | 880 | 703 | 906 | 1166 | 160 | 1207 | 1 | CA |
| 8 | 01_alb_id/images/CA/CA01_13.tif | 204 | 774 | 869 | 752 | 890 | 1169 | 220 | 1200 | 1 | CA |
| 9 | 01_alb_id/images/CA/CA01_16.tif | 146 | 759 | 856 | 737 | 880 | 1180 | 164 | 1217 | 1 | CA |
| 10 | 01_alb_id/images/CA/CA01_28.tif | 96 | 673 | 904 | 641 | 932 | 1145 | 122 | 1192 | 1 | CA |
| 11 | 01_alb_id/images/CA/CA01_24.tif | 57 | 645 | 933 | 614 | 962 | 1154 | 87 | 1205 | 1 | CA |
| 12 | 01_alb_id/images/CA/CA01_15.tif | 165 | 768 | 860 | 746 | 882 | 1177 | 183 | 1212 | 1 | CA |
| 13 | 01_alb_id/images/CA/CA01_17.tif | 137 | 755 | 866 | 730 | 892 | 1185 | 157 | 1225 | 1 | CA |
| 14 | 01_alb_id/images/CA/CA01_12.tif | 210 | 771 | 871 | 749 | 893 | 1164 | 227 | 1195 | 1 | CA |
| 15 | 01_alb_id/images/CA/CA01_21.tif | 102 | 714 | 919 | 688 | 946 | 1196 | 127 | 1241 | 1 | CA |
| 16 | 01_alb_id/images/CA/CA01_29.tif | 101 | 676 | 906 | 647 | 934 | 1151 | 127 | 1195 | 1 | CA |
| 17 | 01_alb_id/images/CA/CA01_10.tif | 177 | 757 | 865 | 732 | 889 | 1163 | 195 | 1199 | 1 | CA |
| 18 | 01_alb_id/images/CA/CA01_04.tif | 98 | 687 | 892 | 656 | 919 | 1149 | 123 | 1195 | 1 | CA |

**Figure 2:** The structure of the converted data

All the concatenate layers implement a kind of skip-connections in the CNN. Despite this decision layer inside the model, it is still a semantic segmentation network that produces a probability map to define whether a pixel belongs to an identity document or not.

For architecture details, data dimensionality, hyper-parameters, and the number of neurons in layers refer to Fig. 3. The optimizer is 'Adam,' Keras built-in. The learning rate is 0.001. The number of training epochs is 60. The loss function is binary cross-entropy. The metrics are the following: accuracy, precision, recall. There are a total of 198,273 trainable model parameters. The size of the model is 832 KiB. It appears to be small enough to run on a smartphone or one-board microcomputer.

We used the TensorFlow [20] and Keras [21] frameworks in our work. The Tensorboard was used for the visualization of training scalars and neural network structures.

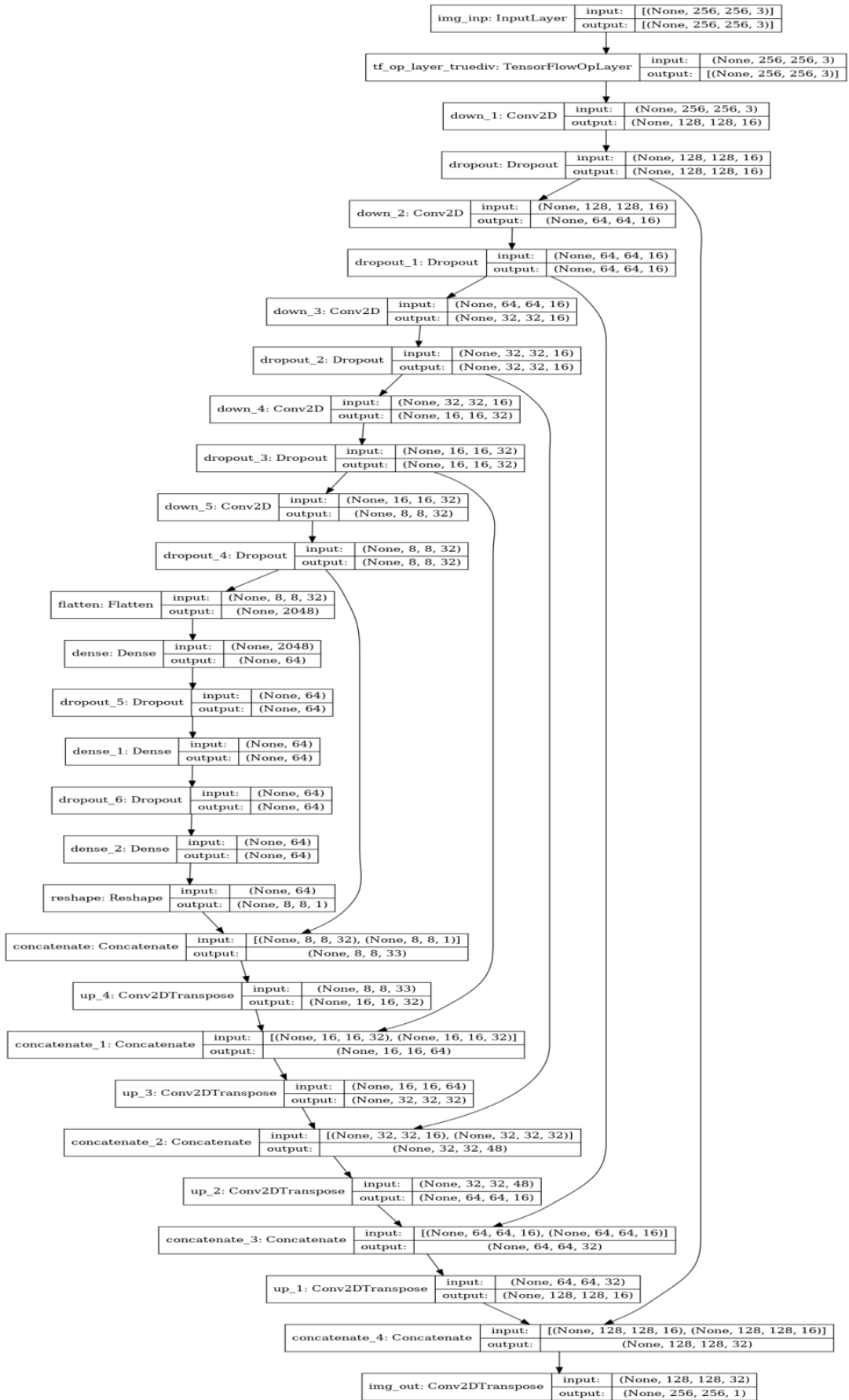

**Figure 3:** The architecture of the proposed CNN

## 5. Training and Evaluation

Training of the model was performed using a conventional server with Intel(R) Core(TM) i7-9700K CPU, 3.60 GHz with 64 GiB of RAM. The training procedure takes approximately 9 ms per one sample, 290 ms per step (batch), 95 seconds per epoch. The number of samples per gradient update (the batch size) is 32. The training and validation loss, accuracy, precision, and recall versus epoch number are presented in Fig. 4 and 5. Values are taken at the end of each epoch.

We used post-predict evaluation in order to evaluate the model. The test set went through the prediction method. After that, predictions were compared to the ground truth and the confusion matrix was derived. The following class-wise metrics were obtained from the confusion matrix: accuracy, true positive rate (TPR, recall), positive predictive value (PPV, precision), etc.

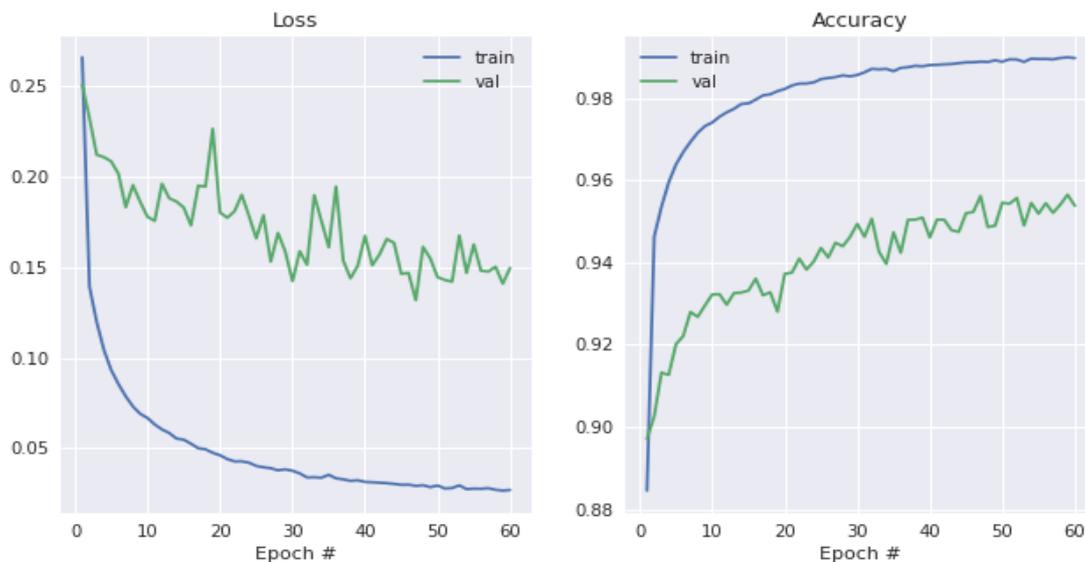

**Figure 4:** Loss and accuracy versus the number of epochs

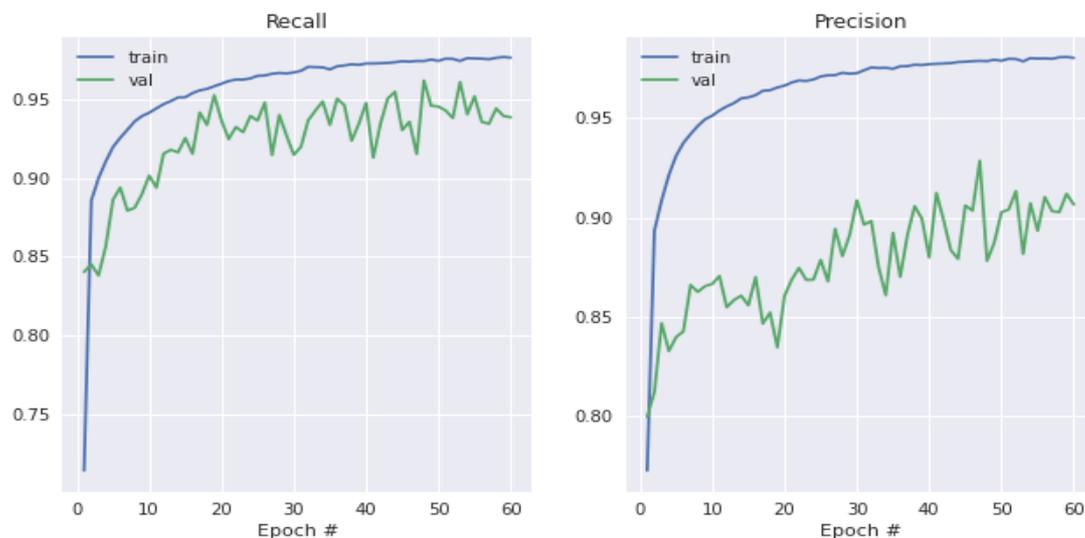

**Figure 5:** Precision and recall versus the number of epochs

## 6. Results

The plot of accuracy versus Intersection over Union (IoU) threshold value is presented in Fig. 6. We achieved an accuracy value of 0.77 for an IoU threshold value of 0.8 on the test set. That is much better in comparison with the simple OpenCV-based approach (accuracy value of 0.32 for this dataset). An example of the model input, ground truth, and prediction is presented in Fig. 7.

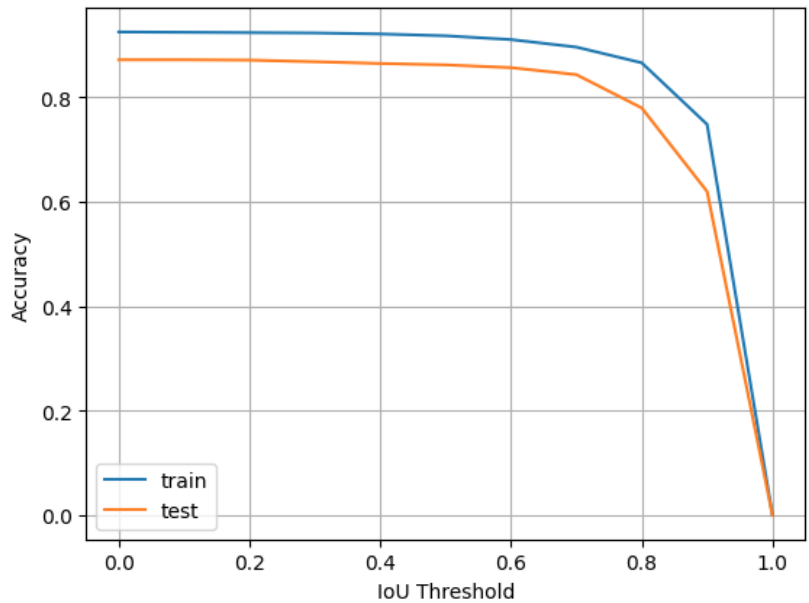

**Figure 6:** Accuracy versus the IoU threshold value

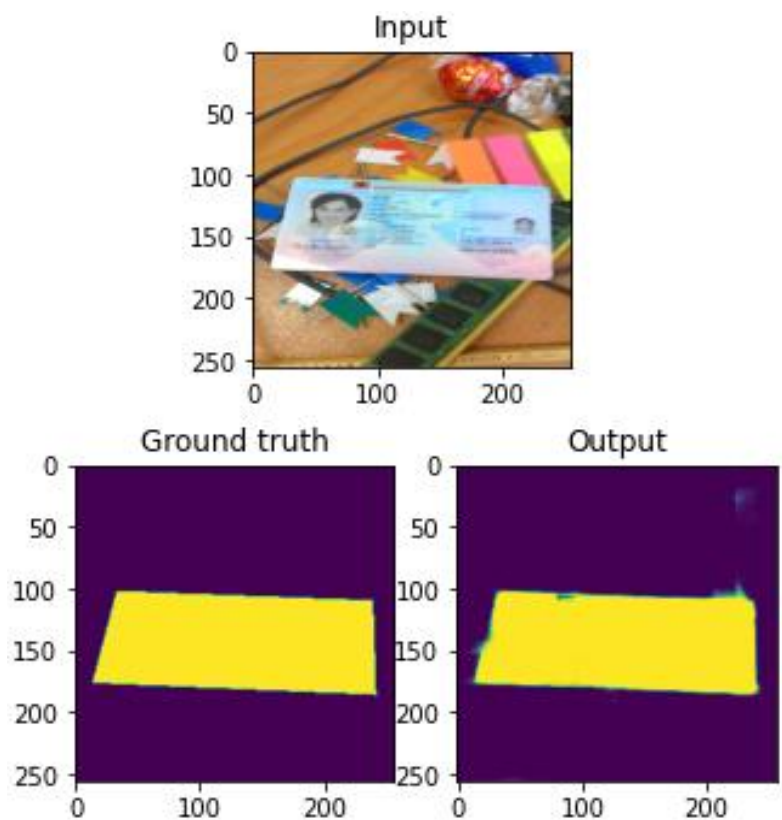

**Figure 7:** Model input, ground truth, and the prediction

After NN makes its prediction on the resize of a given image, we threshold the result by 0.5 and search it for all the contours. After we smooth each contour, we check whether it has four edges and if it occupies the minimum allowed area. If yes, it is checked to be the biggest among other such contours. The selected contour is resized correspondingly to an input image and the rectangle is extracted using OpenCV tools. The result is shown in Fig. 8.

239

Time complexity is one of the most important issues related to real-time data processing. We found the run-time complexity of the detection by measuring the time of one image processing on the needed hardware platform. The average processing time of one image is 8 ms. So, it is possible to perform real-time object detection with the mentioned above hardware platform. Detailed Python3 code of the working prototype we provide in [22].

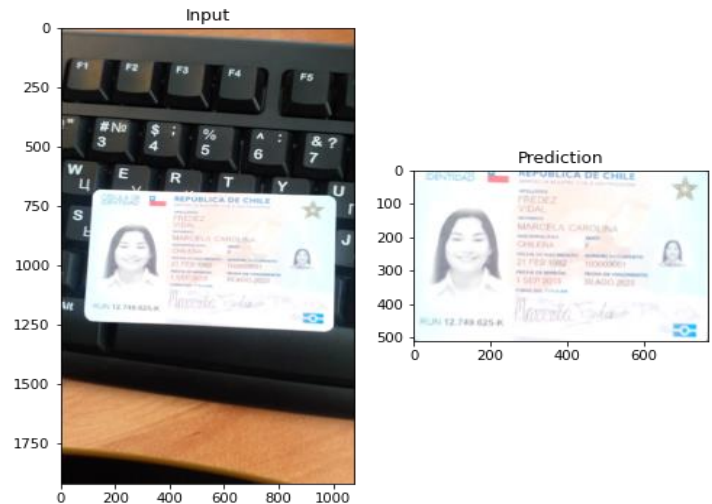

**Figure 8:** The input image and the prediction

## 7. Conclusion and discussion

The overall purpose of the study was to prove the feasibility of efficient identity document detection using the convolution neural network of the proposed architecture. Our main finding suggests that the use of the proposed CNN has an acceptable outcome. CNN layers, as feature extractors, and dense neural layers are easy to implement computational structures with modern hardware platforms such as smartphones, microcontrollers, and industrial one-board microcomputers. They can be easily implemented using modern software frameworks. So, it is possible to build different applications and services using this approach. As stated above, the accuracy of the method is high enough. An important advantage of the proposed method is the ability to permanently retraining on new data. This makes it easy to adapt to new conditions and image properties.

## 8. Limitations and further research

The concern about the study was the limitation of the use of only one dataset. Other data might have different properties. Therefore, there is a need to evaluate the model on other data. In addition, tuning the hyperparameters issue is to be studied. The limitations of the study are not fatal and will be addressed in our future research. Also, we are planning to apply this semantic segmentation-based deep learning approach to process one-dimensional [23] and three-dimensional LiDAR data.

## 9. Acknowledgment

The authors gratefully acknowledge the contributions of scientists of the MindCraft AI LLC and the Department of Information Technology of the Vasyl Stefanyk Precarpathian National University for scientific guidance given in discussions and technical assistance helped in the actual research.

## 10. Disclosures

The authors declare that there is no conflict of interest.